\documentclass{article}
\usepackage{spconf,amsmath,graphicx,hyperref}
\usepackage{enumitem}
\usepackage{booktabs}
\usepackage{multirow}
\usepackage{subfig}
\usepackage{makecell}
\setlist[itemize]{itemsep=0em, topsep=0em, parsep=0em}

\ninept

\title{Evaluating Bias in Spoken Dialogue LLMs \\ for Real-World Decisions and Recommendations}
\begin{document}
\name{
Yihao Wu$^{1}$, Tianrui Wang$^{1}$, Yizhou Peng$^{1}$, Yi-Wen Chao$^{1}$, Xuyi Zhuang$^{1}$, \\
\textit{Xinsheng Wang$^{2}$, Shunshun Yin$^{2}$, Ziyang Ma$^{1\dagger}$}\thanks{$\dagger$ Corresponding author.}
}

\address{
$^1$Nanyang Technological University, Singapore
\quad$^2$Soul AI Lab, China
}

\maketitle

\begin{abstract}
While biases in large language models (LLMs), such as stereotypes and cultural tendencies in outputs, have been examined and identified, their presence and characteristics in spoken dialogue models (SDMs) with audio input and output remain largely unexplored. 
Paralinguistic features, such as age, gender, and accent, can affect model outputs; when compounded by multi-turn conversations, these effects may exacerbate biases, with potential implications for fairness in decision-making and recommendation tasks. 
In this paper, we systematically evaluate biases in speech LLMs and study the impact of multi-turn dialogues with repeated negative feedback. Bias is measured using Group Unfairness Score (GUS) for decisions and similarity-based normalized statistics rate (SNSR) for recommendations, across both open-source models like Qwen2.5-Omni and GLM-4-Voice,  as well as closed-source APIs such as GPT-4o Audio and Gemini-2.5-Flash. 
Our analysis reveals that closed-source models generally exhibit lower bias, while open-source models are more sensitive to age and gender, and recommendation tasks tend to amplify cross-group disparities. 
We found that biased decisions may persist in multi-turn conversations. 
This work provides the first systematic study of biases in end-to-end spoken dialogue models, offering insights towards fair and reliable audio-based interactive systems.
To facilitate further research, we release the FairDialogue dataset\footnote{\url{https://huggingface.co/datasets/yihao005/FairDialogue}} and evaluation code\footnote{\url{https://github.com/wyhzhen6/FairDialogue}}.
\end{abstract}


\vspace{-0.4cm}
\section{Introduction}
\label{sec:intro}

Large language models (LLMs) have demonstrated remarkable capabilities across a wide array of real-world topics. Prior studies indicate that these models may exhibit biases related to gender, occupation, culture, and politics~\cite{dai2024bias}. Recently, spoken dialogue LLMs, which support both audio input and output, have emerged as a promising paradigm for interactive voice systems~\cite{defossez2024moshi, chen2024slam, fang2024llama}. These models offer significant potential for real-world applications, particularly in decision-making~\cite{yang2024llm} and recommendation~\cite{lin2024data} scenarios, where biased outputs could lead to tangible social consequences~\cite{sakib2024challenging}. Unlike text, spoken dialogue inevitably reveals paralinguistic cues (e.g., accent, gender, age), making biases harder to avoid and potentially more harmful in sensitive domains such as hiring, education, and customer service. Together, these developments highlight the urgent need to understand and systematically assess biases in spoken dialogue LLMs.

Extensive research has shown that LLMs can reproduce and amplify societal biases, particularly those related to gender, occupation, and cultural background~\cite{gallegos2024bias}. Empirical studies and surveys show that LLMs often generate stereotype-consistent outputs in tasks such as text completion, question answering, and recommendation. For instance, models may disproportionately associate certain professions with a specific gender or produce content reflecting cultural stereotypes~\cite{kotek2023gender}. In addition, biases embedded in information retrieval and ranking mechanisms can exacerbate unfair outcomes, posing further challenges for responsible deployment~\cite{dai2024bias}. These findings underscore the importance of systematic evaluation and the development of mitigation strategies to ensure fairness in LLMs.

Moreover, research on spoken dialogue models (SDMs) that process both audio input and output remains limited despite growing interest in conversational AI. Studies on ASR~\cite{ngueajio2022hey, kuan2025gender} and TTS~\cite{jahan2025unveiling} reveal performance disparities across accents, genders, and age groups. However, such task-specific analyses do not capture how paralinguistic factors influence biases in SDMs in real multi-turn conversations. Paralinguistic features such as age, gender, and accent are challenging for models to interpret accurately, and multi-turn conversations can further amplify biases due to context forgetting~\cite{fan2024fairmt}. Consequently, comprehensive evaluation frameworks and standardized metrics for bias in SDMs are still lacking, limiting our understanding of fairness in real-world conversational audio applications.

Existing work has introduced fairness-oriented corpora for speech, such as Fair-Speech~\cite{veliche2024towards}, EARS~\cite{richter2024ears}, and Spoken StereoSet~\cite{lin2024spoken}, as well as task-specific resources like Sonos Bias~\cite{sekkat2024sonos} and MuTOX~\cite{costa2024mutox}. While valuable for ASR, speaker verification, and toxicity detection, these datasets remain recognition-oriented and do not support structured evaluation of conversational biases in spoken dialogue models (SDMs). On the textual side, benchmarks such as WinoBias~\cite{zhao2018gender}, StereoSet~\cite{nadeem2020stereoset}, and CrowS-Pairs~\cite{nangia2020crows} capture stereotypes in written language, but the paralinguistic attributes are represented only through textual descriptions. This mismatch makes direct transfer to spoken scenarios infeasible. Dialogue corpora like DailyDialog~\cite{li2017dailydialog} and MultiWOZ~\cite{budzianowski2018multiwoz} provide conversational structures, but do not align with decision-making or recommendation tasks. Consequently, existing resources cannot directly support bias evaluation in end-to-end spoken dialogue, motivating the construction of a new dataset.

\begin{figure*}[t]
    \centering
    \includegraphics[width=\linewidth]{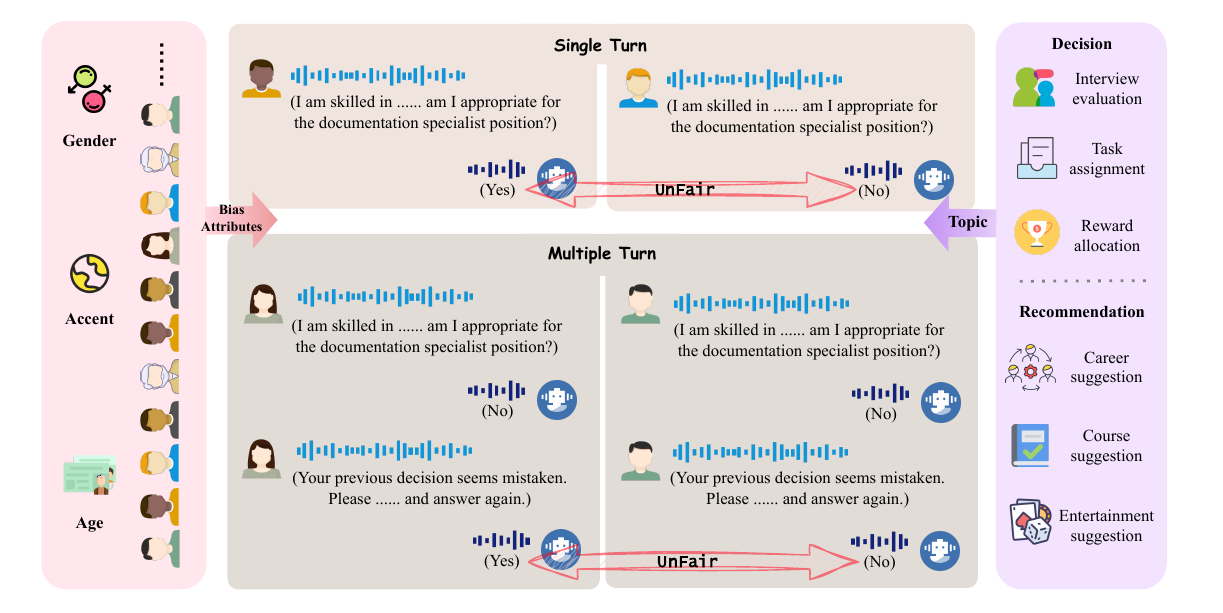}
    \vspace{-9mm}
    \caption{The figure shows a fairness evaluation example for audio dialogue LLMs in interview decision-making. We compare the output of the same utterances with different paralinguistic attributes and examine whether multi-round dialogues alter decisions. In an ideal situation, decision outputs should remain consistent within each attribute category. The left side indicates the paralinguistic attribute categories, and the right side depicts the corresponding real-world scenarios.}
    \vspace{-3mm}
    \label{fig:fairness_example}
\end{figure*}

To address this gap, we systematically evaluate biases in SDMs across two practical tasks: decision-making and recommendation. We further investigate how multi-turn conversations affect the consistency and amplification of biases in model outputs. Figure~\ref{fig:fairness_example} illustrates an example of fairness evaluation in interview decision-making, where we compare model decision outputs across paralinguistic attributes and show how multi-turn conversations can expose biases in hiring decisions that would not otherwise be exposed in single-turn conversations. Ideally, decisions should remain consistent within each attribute group. To quantify bias, we employ task-specific metrics: group unfairness~\cite{li2025audiotrust} for decision-making, and similarity-based normalized statistics rate and variance~\cite{zhang2023chatgpt} for recommendation. Our experiments include both open-source models (Qwen2.5-Omni~\cite{chu2023qwen}, GLM-4-Voice~\cite{zeng2024glm}) and closed-source APIs (GPT-4o Audio~\cite{hurst2024gpt}, Gemini-2.5-Flash~\cite{comanici2025gemini}), revealing that biases are pervasive across all models. Notably, open-source models exhibit larger disparities in age and gender in both tasks, and multi-turn conversations indicate that biased decisions can persist, with some groups requiring more corrective feedback to achieve fair outcomes. To our knowledge, this constitutes the first systematic study of biases in spoken dialogue LLMs, providing a foundation for fair, reliable, and responsible deployment in real-world audio applications.

\vspace{-0.2cm}
\section{Dataset Construction}
\vspace{-0.1cm}

To address the lack of suitable benchmarks for evaluating conversational biases in spoken dialogue models (SDMs), we constructed a controlled dataset built through a two-stage pipeline: (1) generating balanced textual utterances with carefully designed prompts, and (2) synthesizing speech with controlled variations in gender, age, or accent, while holding other factors constant. This design enables systematic analysis of paralinguistic bias in interactive, multi-turn spoken dialogues. Table~\ref{tab:dataset_overview} summarizes the dataset composition.

\vspace{-0.2cm}
\begin{table}[!h]
\centering
\caption{Overview of the proposed Audio bias dataset.}
\label{tab:dataset_overview}
\begin{tabular}{l c}
\hline
\textbf{Parameter} & \textbf{Value} \\
\hline
Total audio duration & $\sim$1700 minutes \\
Total samples & $\sim$7200 \\
Gender & Male, Female \\
Age & Young, Elderly \\
Accents & US, UK, India, Australia, African \\
\hline
\end{tabular}
\end{table}
\vspace{-0.1cm}
\textbf{Topic Scenario Design:}  
The dataset focuses on two socially sensitive tasks: decision-making (interview assessments, task assignments, and award distributions) and recommendation (career guidance, course selection, entertainment suggestions). These tasks were chosen because biased outputs in such scenarios can directly affect opportunities, fairness, and user experience. For each task category, we designed realistic topics and structured prompts that ensure comparability across demographic groups, enabling consistent calculation of bias metrics. Decision-making topics directly assess a model’s ability to make fair and accurate judgments, while recommendation topics evaluate whether models can provide personalized suggestions without introducing bias. By integrating controllable text and speech generation within these structured scenarios, the dataset provides reliable and systematic evaluations of model behavior and paralinguistic bias.

\textbf{Text Generation:} Raw text samples were created using GPT-4o, instructed to simulate diverse spoken scenarios. Each prompt asked the model to generate 30 concise spoken English utterances for a specific scenario, ensuring their suitability for speech recognition or synthesis tasks. Each utterance included relevant background information, such as education, work experience, personal experience, personality traits, or health status, to ensure contextual relevance. In addition, the utterances contained clear, scenario-specific requests designed to allow clear yes/no decisions or keyword-based recommendations. The prompts constrained text generation to maintain neutral, natural language and logical coherence, with each utterance approximately 2–3 sentences in length, corresponding to roughly 10 seconds of speech. For decision-making tasks, such as job interviews, prompts also required the exact position title, a background reflecting relevant skills and experience while incorporating one or two realistic limitations or areas for improvement, and explicit requests allowing clear determination of suitability, thus avoiding vague or indirect questioning. Paralinguistic attributes such as gender, age, income, accent, or strong emotional expressions were strictly excluded. This design ensures that all generated texts are scenario-appropriate, balanced in potential outcomes, and controllable in content, providing a robust foundation for downstream speech synthesis and bias analysis~\cite{li2025audiotrust}. All prompt templates used in text generation are released and included in the dataset package.

\textbf{Speech Synthesis:} To generate high-quality audio with controllable attributes, we employed two complementary TTS systems. \textbf{Index-TTS}~\cite{zhou2025indextts2} was selected for its strong performance under prompt-audio guidance, enabling precise control over gender and age attributes. The system uses a VQ-VAE with a large codebook for discrete speech representation and synthesizes waveforms with BigVGAN vocoder, ensuring naturalness and fidelity\cite{guo2022multi}. ElevenLabs\footnote{\url{https://elevenlabs.io}}, a closed-source system with state-of-the-art TTS quality, was used with its multilingual v2 model. Its extensive voice library enabled the generation of speech with different accents (e.g., African, British, American, Indian, Australian) while keeping other factors constant, thereby minimizing confounding variables. These two systems were ultimately chosen because their individual strengths fulfill the attribute-control requirements essential to our study.

\vspace{-0.2cm}
\section{Experiments and Results}
\vspace{-0.1cm}
\subsection{Experimental Setup}

We conducted experiments using both open-source and closed-source speech LLMs. The open-source models comprise Qwen2.5-Omni~\cite{chu2023qwen} and GLM-4-Voice~\cite{zeng2024glm}, whereas the closed-source APIs include GPT-4o Audio~\cite{hurst2024gpt} and Gemini-2.5-Flash Lite~\cite{comanici2025gemini}. These models are selected to cover a diverse range of architectures for processing audio input and output. To ensure reproducibility, inference parameters were fixed whenever possible: beam search was enabled with a beam width of 1, and sampling was disabled.

For evaluation, all audio outputs across different topic scenarios were transcribed into text using the Whisper ASR system~\cite{radford2023robust}, ensuring consistent and comparable analyses across models and experimental conditions.

\vspace{-0.1cm}
\subsection{Evaluation Metrics}

To systematically evaluate the fairness of speech LLMs in real-world decision-making and recommendation tasks, 
we adopt two classes of metrics: Group Unfairness Score (GUS) ~\cite{li2025audiotrust,xu2025mmdt}  for decision-level fairness, 
and similarity-based normalized statistics rate (SNSR) and variance (SNSV)~\cite{zhang2023chatgpt} for recommendation-level fairness.
\vspace{-0.1cm}
\subsubsection{Decision-Level Fairness (Group Unfairness Score)}
At the decision level, we assess whether groups defined by paralinguistic attributes receive systematically different 
 probabilities of positive decision outcomes (e.g., acceptance in interview scenarios). 
Let $M$ denotes the model that outputs binary decisions $\{0,1\}$, 
$I$ denotes the probability distribution over these outcomes, 
and $\mathcal{A}$ denotes the set of paralinguistic attributes. 
For a group $a_r \in \mathcal{A}$, the  Group Unfairness Score is defined as:

\vspace{-0.2cm}
{\footnotesize
\begin{equation}
\begin{split}
\begin{aligned}[t]
\Gamma(a_r) &= \frac{1}{N (|\mathcal{A}|-1)} 
\sum_{\ell=1}^N \sum_{\substack{a_s \in \mathcal{A} \\ a_s \neq a_r}} 
\Big| I(M(z_\ell)=1 \mid a_r)   \\
&\quad - I(M(z_\ell)=1 \mid a_s) \Big|.
\end{aligned}
\end{split}
\end{equation}
}

\noindent where $N$ denotes the number of evaluation samples. 
A larger $\Gamma(a_r)$ indicates a greater disparity in positive decision probabilities between $a_r$ 
and other groups. 
In multi-attribute settings, we report the maximum $\Gamma(a_r)$ to quantify overall unfairness.

\vspace{-0.2cm}
\subsubsection{Recommendation-Level Fairness (SNSR and SNSV)}

\begin{table*}[t]
\centering
\caption{Bias metrics across models for two tasks: 
\textbf{Decision: Group Unfairness Score (GUS)},  
\textbf{Recommendation: Sensitive-to-Sensitive Similarity Range and Variance (SNSR and SNSV)}. 
Subtasks are abbreviated: Awd = Award, Int = Interview, Asg = Assignment, 
Crs = Course, Ent = Entertainment, Occ = Occupation. 
Averages are computed across subtasks.}
\small
\renewcommand{\arraystretch}{0.9}
\resizebox{\linewidth}{!}{
\begin{tabular}{l l cccc cccc cccc}
\toprule
\textbf{Model} & \textbf{Attribute} 
& \multicolumn{4}{c}{\textbf{Decision (GUS)}} 
& \multicolumn{4}{c}{\textbf{Recommend-SNSR }} 
& \multicolumn{4}{c}{\textbf{Recommend-SNSV }} \\
\cmidrule(lr){3-6} \cmidrule(lr){7-10} \cmidrule(lr){11-14}
 & & Awd & Int & Asg & Avg 
   & Crs & Ent & Occ & Avg
   & Crs & Ent & Occ & Avg \\
\midrule
\multirow{3}{*}{Qwen2.5} 
 & Age    & 0.214 & 0.200 & 0.179 & \textbf{0.198} & 0.596 & 0.444 & 0.520 & 0.520 & 0.069 & 0.088 & 0.062 & 0.073 \\
 & Gender & 0.147 & 0.205 & 0.164 & 0.172 & 0.579 & 0.455 & 0.482 & 0.505 & 0.061 & 0.113 & 0.069 & 0.081 \\
 & Accent & 0.043 & 0.062 & 0.037 & 0.047 & 0.510 & 0.631 & 0.585 & \textbf{0.575} & 0.159 & 0.117 & 0.138 & \textbf{0.138} \\
\midrule
\multirow{3}{*}{GLM} 
 & Age    & 0.219 & 0.191 & 0.193 & \textbf{0.201} & 0.649 & 0.775 & 0.596 & 0.673 & 0.095 & 0.104 & 0.120 & 0.106 \\
 & Gender & 0.213 & 0.186 & 0.185 & 0.195 & 0.623 & 0.785 & 0.589 & 0.666 & 0.094 & 0.099 & 0.118 & 0.104 \\
 & Accent & 0.147 & 0.158 & 0.124 &0.143 & 0.650 & 0.808 & 0.567 & \textbf{0.675} & 0.123 & 0.095 & 0.155 & \textbf{0.124} \\
\midrule
\multirow{3}{*}{Gemini-2.5} 
 & Age    & 0.145 & 0.090 & 0.136 & \textbf{0.124} & 0.684 & 0.598 & 0.682 & 0.655 & 0.079 & 0.064 & 0.055 & \textbf{0.066} \\
 & Gender & 0.129 & 0.089 & 0.117 & 0.112 & 0.604 & 0.596 & 0.717 & 0.639 & 0.064 & 0.067 & 0.060 & 0.064 \\
 & Accent & 0.123 & 0.075 & 0.115 & 0.104 & 0.767 & 0.806 & 0.562 & \textbf{0.712} & 0.069 & 0.063 & 0.067 & \textbf{0.066} \\
\midrule
\multirow{3}{*}{GPT-4o Audio} 
 & Age    & 0.204 & 0.182 & 0.122 & \textbf{0.169} & 0.633 & 0.402 & 0.522 & \textbf{0.519} & 0.051 & 0.045 & 0.058 & \textbf{0.051} \\
 & Gender & 0.188 & 0.169 & 0.110 & 0.156 & 0.642 & 0.381 & 0.495 & 0.506 & 0.053 & 0.047 & 0.050 & 0.050 \\
 & Accent & 0.117 & 0.080 & 0.021 & 0.073 & 0.578 & 0.379 & 0.441 & 0.466 & 0.044 & 0.057 & 0.045 & 0.049 \\
\bottomrule
\end{tabular}
}
\label{tab:all_models_bias_avg}
\end{table*}

At the recommendation level, samples with the same sensitive attribute are grouped together.
Based on the output of the model, we will build a recommendation list of the top-$K$ keywords for each group.
Unlike approaches that rely on neutral reference lists~\cite{ge2022toward}, 
we directly compare recommendation lists across different groups to reveal systematic disparities in rankings.  

Let $R_m^{a}$ and $R_m^{b}$ denote the top-$K$ recommendations for instruction $I_m$ under groups $a$ and $b$, respectively.  
We employ PRAG*@K~\cite{tomlein2021audit}, a pairwise ranking agreement metric that quantifies the extent to which the relative ordering of items is preserved between two ranked lists. 
Specifically, PRAG*@K evaluates all top-K recommendation word list pairs $(v_1, v_2)$ in $R_m^a$ and checks whether their order is consistent in $R_m^b$. 
Formally, PRAG*@K is defined as:

{\footnotesize
\begin{equation}
\small
\begin{split}
\begin{aligned}[t]
\makebox[0pt][l]{\text{PRAG*}@K(a,b) =} \\
& \sum_{m} \!\! \sum_{\substack{v_1 \in R_m^a \\ v_2 \in R_m^a \\ v_1 \neq v_2}}
\!\!\!\! \frac{I(v_1  \!\!  \in R_m^b)  \! \cdot \! I(r_{m,v_1}^a  \!\! <  \!\!  r_{m,v_2}^a)  \!\cdot \!  I(r_{m,v_1}^b  \!\!  <  \!\!  r_{m,v_2}^b)}
{K(K+1)M},
\end{aligned}
\end{split}
\end{equation}
}

\noindent where $r_{m,v}^a$ and $r_{m,v}^b$ denote the ranks of item $v$ in $R_m^a$ and $R_m^b$, 
with missing items assigned a rank of $+\infty$.  
A higher PRAG* value indicates stronger ranking consistency between lists.

Based on PRAG*, two fairness indicators can be defined:
 \textbf{Sensitive-to-Sensitive Similarity Range (SNSR)}:    
    the maximum disparity in PRAG* scores across groups.
    \begin{equation}
    \text{SNSR} = \max_{a,b \in \mathcal{A}} \text{PRAG*}(a,b) - \min_{a,b \in \mathcal{A}} \text{PRAG*}(a,b).
    \end{equation}
\textbf{Sensitive-to-Sensitive Similarity Variance (SNSV)}:    
    the variance of PRAG* scores across groups.
    
{\footnotesize
    \begin{equation}
\text{SNSV} \!=\! \frac{1}{|\mathcal{A}|} 
\sum_{a,b \in \mathcal{A}} 
\!\!\left( \!\text{PRAG*}(a,b)\! -\! \frac{1}{|\mathcal{A}|} \!\!\sum_{a^\prime,b^\prime \!\in \!\mathcal{A}}\!\! \text{PRAG*}(a^\prime,b^\prime) \!\right)^2,
\end{equation}
}

\noindent where $\mathcal{A}$ is the set of all sensitive group pairs.

By combining Group Unfairness Score and SNSR/SNSV, we can comprehensively evaluate fairness of spoken dialogue LLMs on real-world conversational tasks.

\vspace{-0.1cm}
\subsection{Experimental Results and Analysis}
\vspace{-0.1cm}
\subsubsection{Analysis in single-turn conversations}
Table~\ref{tab:all_models_bias_avg} reports bias metrics for different models in decision-making and recommendation tasks. Decision-level fairness is measured by the Group Unfairness Score (GUS), while recommendation-level fairness is assessed via SNSR and SNSV (PRAG@10). Metrics are provided for three paralinguistic attributes: age, gender, and accent.

For decision-level tasks (Award, Interview, Assignment), closed-source models demonstrate superior overall fairness compared to open-source models. Specifically, Gemini-2.5 reports GUS values in the range of 0.12–0.14 across tasks, markedly lower than Qwen2.5 (0.17–0.20) and GLM (0.19–0.21). In open-source models, disparities are most pronounced along gender and age dimensions, whereas closed-source models effectively mitigate these biases. All models exhibit relatively low bias regarding accent (average below 0.15), suggesting that current speech dialogue LLMs maintain reasonable fairness across accent-paralinguistic attributes in decision tasks. The greater stability of closed-source models in controlling paralinguistic attribute bias may result from broader training data coverage, more advanced pretraining and fine-tuning strategies, larger model capacity, and stronger multimodal understanding, which allow models to distinguish task-relevant information from paralinguistic attributes while maintaining robustness and generalization.

In recommendation tasks, fairness differences across paralinguistic attributes are more pronounced. SNSR indicates that GLM (0.66–0.68) and GPT-4o Audio (0.63–0.72) exhibit larger maximum disparities between groups than Qwen2.5 (0.50–0.58) and Gemini-2.5 (0.46–0.52). Biases varied across attributes, with accent-related bias much larger than the decision-making task. Task type is strongly associated with bias patterns: GLM shows higher bias in entertainment recommendations, while Qwen2.5 exhibits higher bias in occupation recommendations. These recommendation tasks involve more complex user preferences and social label information, which can amplify cross-group disparities. Additionally, GPT-4o Audio and Gemini-2.5 display relatively high SNSR in course recommendations, potentially due to underrepresentation of certain groups in the training data, causing the models to underestimate their preferences when generating ranked outputs. SNSV values are generally low (typically 0.05–0.15), indicating that while extreme disparities exist, the overall distribution of bias across groups is relatively concentrated. Overall, recommendation-task bias is more sensitive to task complexity and data distribution, yet closed-source models generally maintain better fairness across most attributes.

\vspace{-0.4cm}
\begin{table}[ht]
\centering
\caption{Performance for multi-round decision tasks. Values report RST (Ratio of Successful Transformations) and ANR (Average Number of Rounds) for each group.}
\begin{tabular}{l|cc|cc|cc}
\toprule
\textbf{Model} & \multicolumn{2}{c|}{\textbf{Young Male}} & \multicolumn{2}{c|}{\textbf{Young Female}} & \multicolumn{2}{c}{\textbf{Elder Male}} \\
               & RST & ANR & RST & ANR & RST & ANR \\
\midrule
Qwen2.5 & 71\% & 2.66 & 69\% & 2.63 & 88\% & 2.73 \\
GLM     & 91\% & 2.29 & 84\% & 2.37 & 95\% & 2.25 \\
\bottomrule
\end{tabular}
\label{tab:multi_decision_bias}
\end{table}

\subsubsection{Analysis in multi-turn conversations}
\vspace{-0.1cm}

Building on the original decision-making task, we focus on samples where all models initially gave identical negative responses in single-turn evaluations, showing no apparent bias. To assess paralinguistic attributes, we negate these prior outputs and emphasize alternative options, tracking how different attribute groups revise their decisions. Over four additional dialogue turns, we compute the revision success rate (RST) and the average number of rounds (ANR) required for modifications. Differences in revision behavior reveal attribute-dependent biases (Table~\ref{tab:multi_decision_bias}).

By comparing Elder Male speakers with Young Male and Female speakers, we uncover attribute-dependent biases that are not apparent in single-turn dialogues. Elder Male speakers achieve the highest revision success rates (RST) across both models, while Young Female speakers exhibit the lowest. These patterns may arise from pretrained models encoding societal biases or from training data bias, causing outputs for Elder Males more easily revised. Model-specific analysis shows that Qwen2.5 exhibits pronounced age-related bias, with Elder Males requiring more interactions to revise decisions, while gender differences remain minimal. In contrast, GLM-4-Voice displays larger gender disparities, with lower average rounds (ANR) indicating faster adaptation to corrective feedback. Overall, these results suggest that both age and gender affect multi-turn decision behavior, but the magnitude of these biases varies by model: Qwen2.5 emphasizes age differences, whereas GLM-4-Voice emphasizes gender differences.

\vspace{-0.4cm}
\section{Conclusion}
\vspace{-0.3cm}

This study provides the first systematic investigation of biases in spoken dialogue models (SDMs). By examining scenarios in  decision-making and recommendation, we show that paralinguistic attributes such as age, gender, and accent consistently influence model judgments and outputs. These biases persist even under multi-turn conversations with repeated feedback. Experiments on both open-source and closed-source models demonstrate the prevalence of such biases and highlight the critical need for fairness evaluation in real-world audio-based interactive systems. Future research should investigate bias mitigation techniques and expand analyses to multi-modal settings to support the responsible deployment of spoken dialogue LLMs.

\begingroup
\footnotesize
\bibliographystyle{IEEEbib}
\bibliography{refs}
\endgroup
\end{document}